# TweetDIS: A Large Twitter Dataset for Natural Disasters Built using Weak Supervision

**Ramya Tekumalla**
Department of Computer Science
Georgia State University
Atlanta, GA
rtekumalla1@gsu.edu

**Juan M. Banda**
Department of Computer Science
Georgia State University
Atlanta, GA
jbanda@gsu.edu

## Abstract

Social media is often utilized as a lifeline for communication during natural disasters. Traditionally, natural disaster tweets are filtered from the Twitter stream using the name of the natural disaster and the filtered tweets are sent for human annotation. The process of human annotation to create labeled sets for machine learning models is laborious, time consuming, at times inaccurate, and more importantly not scalable in terms of size and real-time use. In this work, we curate a silver standard dataset using weak supervision. In order to validate its utility, we train machine learning models on the weakly supervised data to identify three different types of natural disasters i.e earthquakes, hurricanes and floods. Our results demonstrate that models trained on the silver standard dataset achieved performance greater than 90% when classifying a manually curated, gold-standard dataset. To enable reproducible research and additional downstream utility, we release the silver standard dataset for the scientific community.

## 1 Introduction

Twitter has been an active source of communication, especially during many to many crisis events, such as natural disasters like earthquakes, floods, typhoons, and hurricanes [1]. A wide range of information is tweeted during a disaster by people who are in need of help (e.g., food, shelter, medical assistance, etc.) or by people who are willing to donate or offer volunteering services or by the government to inform people of the latest updates [2] [3]. While a tremendous amount of information is available on social media, the data is massive, noisy, highly distributed, unstructured and dynamic [4]. Hence, it is essential to identify valuable and relevant information from the sea of information [5].

Several studies in the past have demonstrated the role of machine learning in analyzing natural disasters. Ofli et al. [6] utilized machine learning to make sense of aerial data during disasters. Resch et al. [7] utilized topic modeling and spatio temporal analysis of social media data for disaster footprint and damage assessment. Several NLP techniques have been developed to detect and extract relevant information [8] [9]. Nguyen et al. [10] utilized convolutional neural networks to classify crisis related data on social networks. Madichetty et al. [11] demonstrated that contextual representations improve supervised learning when using Twitter data for natural disasters. To encourage reproducible research, several researchers released their benchmark datasets and annotated datasets to the scientific community. However, Twitter users often delete their tweets, causing loss of data and wasted annotation effort unless the original authors save a copy. Generating large training data through manual annotation is impractical since the process is very expensive and time consuming. Thus, there is a huge need to avoid reliance on manual annotation of small datasets and to move to automatic annotation on large datasets.

In this aspect, weak supervision utilizes noisy or imprecise sources to provide a supervision signal for labeling large amounts of training data in a supervised learning setting [12]. By adopting heuristics and defining labeling functions which utilize the heuristics to label the dataset, large training sets i.e silver standard data could be programmatically generated and could be utilized for training machine learning models. The assumption behind our work is that the large volume of training data which can be collected using an automated labeling process, can compensate for the inaccuracy in the labels. Our assumption is based on the theory of noisy learning [13] [14]. In the past, we utilized the weak supervision approach to identify drug mentions on Twitter and determined that the models trained with silver standard data perform similar to the models trained with gold standard data [15]. In this work, we build a heuristic to curate silver standard data and use the silver standard data to train and fine tune several machine learning models. We collected 7,157,153,298 ( 7 billion) tweets and filtered 846,927 natural disaster tweets using a heuristic. Overall, these tasks took around 350 days for hydrating, filtering and classification. This work requires extensive computing power with mass storage options since we had to download, hydrate and store billions of data and the work was computed on a research server with the following configuration: 2 x Intel Xeon-Gold 6148 with 2.4GHz/20-cores, 768 GB RAM, 14.4TB HDD and 7 NVIDIA Tesla V100 GPUs. To encourage reproducible research, we released our silver standard dataset and code to obtain silver standard dataset via Zenodo and the details are mentioned in Section 5.

## 2 Data Preparation

Twitter hosts 187 million users and generates 500 million tweets every day on average [16], making it an attractive social media platform to obtain data for research. In this work, we obtained Twitter data from several sources and used a heuristic to filter relevant tweets. The process of data collection and curation of the heuristic are discussed below in the following sections.

### 2.1 Heuristic Creation

Our objective in this work is to demonstrate the viability of creating and utilizing a silver standard dataset to train machine learning models for an application. We demonstrate that the models trained on silver standard dataset efficiently identify a gold standard dataset which affirms that silver standard dataset can be utilized for training machine learning models. The heuristic is the key to identify the silver standard from a large set of data. We wanted to create a simple heuristic over utilizing labelling functions (Eg: Snorkel framework [17]) as these functions get complicated and are less intuitive for non-specialist user. Since we wanted to curate a silver standard dataset which can be utilized for different kinds of natural disasters and not limit to specific disasters, we did not use the names of the natural disasters (Eg: hurricane Harvey, hurricane Irma, Nepal floods) in our heuristic. In this work we created a heuristic which contains signals from three different natural disasters which are Hurricanes, Earthquakes and Floods. We collected data from previously released datasets for natural disasters and generated bigrams and trigrams to identify strong signals which can be used to describe a natural disaster. To generate bi and trigrams, we preprocessed the tweet text to remove emojis, emoticons, stopwords and lowercased the text. In the following sections, we present the curation process for each kind of natural disaster.

**Hurricanes** Hurricanes have been studied most commonly for research especially for text content analysis and multimedia content analysis [18]. Several datasets have been released in the past for hurricanes which are collected using a keyword based search from Twitter. This results in datasets with additional noise. Hence a heuristic is vital since it can be utilized to filter relevant tweets. We hydrated 4 different publicly available hurricane datasets [19, 20, 21, 22**?** ] and obtained 51,500,116 tweets. We filtered the tweets and obtained only 9,789,940 clean tweets (i.e tweets which are original and not retweeted). We removed terms in the format hurricane <Name> / <Name> hurricane / <hurricane name> term (Example: irma relief). We generated bigrams and trigrams for the clean tweets. An initial analysis on the terms presented overlaps between bigrams and trigrams and hence we used only bigrams to obtain the list of terms. Once we generated the bigrams, we sorted the terms in descending order of the counts and retained the top 150 terms. Further we cleaned up the terms and obtained 62 terms for hurricanes. Example ("hurricane victims", "power outages", "heavy rain").

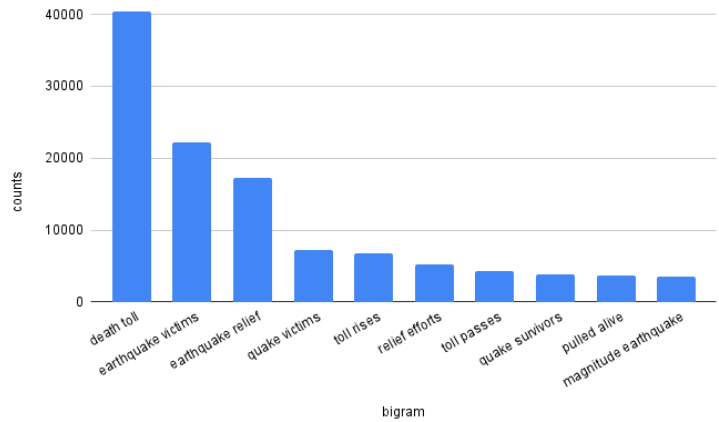

Figure 1: Top 10 frequent bi-grams for 3 different earthquakes.

**Earthquakes and Floods** We did not find any exclusive datasets publicly available for floods and earthquakes. Hence, we compiled the list of floods and earthquakes that occurred between 2018 and 2020 and extracted all relevant tweets from our longitudinal collection of Twitter data. The details of the tweets are presented in the Section 2.2. We included 24 different floods and 4 different earthquakes which occured between 2018 and 2020 to obtain relevant tweets and further generated the bigrams. We sorted the terms in descending order of the counts and retained the top 150 terms. Our final list of terms contain 58 unique flood terms and 48 unique earthquake terms. We eliminated terms which contain the format <Country Name> floods, floods <Country Name>, <Country Name> earthquake. Figure 1 depicts the top 10 most frequent bigrams for earthquakes after filtering the country names.

The primary reason to eliminate the terms which are tied to a particular country (Eg: <country name> floods) is to remove all the terms that can identify one specific event of a disaster. The curation of signals which can identify the natural disaster is what we aimed for. This enhances generalizability when using the heuristic for future natural disasters. Table 1 presents the details of the natural disasters and the number of terms for each natural disaster. There is an overlap (Eg: "death toll") in terms between the three different types of natural disasters. Since we wanted to identify the terms relevant to natural disasters, we filtered the terms and removed duplicates. Our final heuristic contains 155 unique terms.

### 2.2 Collection from Twitter Stream

Using our in-house Social Media Mining Toolkit (SMMT) [23] , we set up a stream collection from Twitter which acquires 1% random sample of tweets from Twitter. This is an ongoing collection which started in 2018. The 1% sample is a sample of the tweets tweeted daily and with Twitter's end point we can only obtain 1% sample. In this work, we used Tweets collected between 2018 and 2021. Table 2 lists the details of tweets collected and filtered from the Twitter Stream. We used only clean English tweets from this stream.

Table 1: No of terms for obtained for each natural disaster

| Natural Disaster | Events Included | Number of terms |
|---|---|---|
| Hurricane | Maria, Sandy, Irma, Harvey | 62 |
| Floods | Rwanda, Kenya, Somalia, Burundi, Djibouti, Ethiopia, Uganda, Japan, Kerala, Vietnam, India, Indonesia, European, Spain, France, Italy, United Kingdom, Portugal, Maryland, Townsville, Venice, Thailand, Pakistan, Iran | 58 |
| Earthquakes | Indonesia, Albania, Fiji, Peru | 48 |

Table 2: Stream collection details

| Year | #tweets | #clean tweets | filtered Tweets |
|---|---|---|---|
| 2018 | 936,487,968 | 45,5783,507 | 108,292 |
| 2019 | 1,180,731,480 | 570,157,502 | 164,933 |
| 2020 | 151,260,4381 | 777,863,405 | 187,401 |
| 2021 (Jan - May) | 621,615,285 | 325,579,195 | 68,762 |
| Total | 4,251,439,114 | 2,129,383,609 | 529,388 |

## 2.3 Publicly available Twitter Datasets

Twitter is heavily used as a data source in many studies to analyze and identify patterns. We identified 34 studies which not only utilized Twitter as their primary source of data but also made their data publicly available enabling reproducible research. These datasets are valuable since they have historic data which are very difficult to obtain. We intend to build a longitudinal dataset which contains tweets from the past. A huge advantage in using the past data is the ability to identfy the shift or trends in dealing with a natural disaster. Commodities that are required during a natural disaster can be easily identified with the past data. Additionally, valuable information like caution and advice can be extracted and can be used for a new natural disaster. The primary intent to use publicly available datasets is to re-use existing work and demonstrate an approach on how existing work can be utilized to build a superior dataset. Further, we observed that data augmentation improves machine learning models when we utilized a heuristic to obtain more relevant tweets to characterize Anti-Asian tweets during Covid-19 pandemic [24]. Hence we utilized publicly available datasets and used our heuristic to filter relevant tweets from the datasets and added the tweets to our silver standard dataset. Since tweet texts cannot be shared publicly, all the studies released the tweet ids which correspond to the tweets they have utilized in their study. The get_metadata utility of the SMMT toolkit was employed to hydrate the tweet ids. The following table summarizes the details of tweets we hydrated using publicly available datasets. A Twitter developer account is required to hydrate the tweets and with a research account, we can hydrate 8,640,000 tweet ids per day. We hydrated a total of 2,905,714,18 (2 billion) tweets in 336 days. We used only clean English tweets for this work. Table 3 presents the details of all the publicly available datasets and number of tweets filtered. All the datasets in the Table 3 are collected based on keyword based search and contain significant noise. The time column in Table 3 represents the time taken (in days) to hydrate each dataset. The filtered tweets column represents the number of tweets filtered using the heuristic.

## 2.4 Gold Standard Dataset Preparation

To test the machine learning models, a gold standard dataset is required. In this work, we utilized a publicly available gold standard dataset [2] which was released in 2016. This publicly available labeled dataset contains data labeled by paid workers [53] and volunteers for several natural disasters like hurricanes, earthquakes, floods, typhoons, landslides. In this work, we utilized their data labeled by paid workers to maintain uniform standards. We also utilized the data for three different types of natural disasters, i.e hurricanes, floods and earthquakes. One of the 9 different labels were available for each tweet in the dataset. **Injured or dead people** indicate reports of casualties and/or injured people due to the crisis. **Missing, trapped, or found people** indicate reports and or questions about missing or found people. **Displaced people and evacuations** indicate information about people who have relocated due to the crisis, even for a short time (includes evacuations). **Infrastructure and utilities damage** indicate reports of damaged buildings, roads, bridges, or utilities/services interrupted or restored. **Donation needs or offers or volunteering services** indicate reports of urgent needs or donations of shelter and/or supplies such as food, water, clothing, money, medical supplies or blood; and volunteering services. **Caution and advice** contain reports of warnings issued or lifted, guidance and tips. **Sympathy and emotional support** indicate prayers, thoughts, and emotional support. **Other useful information** indicates other useful information that helps understand the situation and **not related or irrelevant** indicate unrelated to the situation or irrelevant. We did not use tweets labeled with Donation needs or offers or volunteering services, Sympathy and emotional support and Other useful information in our gold standard dataset as they do not provide any strong signals describing a natural disaster. Tweets labeled as "Not related" are used as negative sets (label 0) for our machine learning models. We cleaned up the gold standard dataset by removing retweets and incomplete tweets. Post clean up, we found that the dataset was imbalanced, hence to balance the dataset, we added few negative tweets to the dataset. The tweets which do not match with any of the patterns in our heuristic were added as negative tweets. A total of **5,692** tweets are used in the gold standard dataset with 2,846 tweets labeled as positive (label 1) and 2,846 labeled as negative (label 0). We would like to emphasize that we did not manually annotate the dataset and we did not pay for the annotate dataset. We used only publicly available, manually annotated datasets released in the past for this work.

Table 3: Details of hydrated datasets

| Dataset | Total Tweet Ids | Total hydrated ids | Uniq Tweet Ids | #clean tweets | Time (in days) | #filtered tweets |
|---|---|---|---|---|---|---|
| 2016 presidential election [25] | 283,244,653 | 122,799,810 | 266,689,265 | 50,788,341 | 33 | 100,998 |
| Solar Eclipse [26] | 13,816,206 | 8,345,117 | 13,763,243 | 1,537,247 | 2 | 639 |
| Hurricane Harvey [19] | 18,352,142 | 10,406,538 | 18,320,786 | 2,142,577 | 2 | 141,635 |
| Hurricane Florence [21] | 7,766,964 | 4,891,342 | 7,747,107 | 1,394,576 | 1 | 94,684 |
| Hurricane Florence [22] | 4,971,575 | 3,399,192 | 4,971,575 | 744,050 | 1 | 62,127 |
| Hurricane Harvey [20] | 7,041,866 | 4,433,003 | 7,041,866 | 883,466 | 1 | 55,287 |
| Hurricane Irma [19] | 17,244,139 | 9,474,907 | 17,212,112 | 2,341,596 | 2 | 79,771 |
| Hurricane Maria [27] | 987,938 | 647,001 | 987,384 | 160,947 | 0 | 5,992 |
| Hurricane Sandy [28] | 14,915,897 | 8,101,431 | 14,913,037 | 5,144,820 | 2 | 177,811 |
| Hurricane Dorian [29] | 3,000,553 | 2,234,048 | 3,000,553 | 416,410 | 0 | 35,099 |
| Hurricane Dorian [30] | 9,186,117 | 6,549,744 | 9,186,117 | 1,723,639 | 1 | 108,883 |
| Election 2012 [31] | 38,393,134 | 22,703,483 | 38,393,134 | 21,751,070 | 4 | 84,084 |
| Datarelease [32] | 106,116,957 | 38,912,028 | 76,187,310 | 30,799,490 | 12 | 16,671 |
| Beyond the Hashtag [33] | 40,815,855 | 23,137,993 | 40,815,855 | 7,307,037 | 5 | 3,344 |
| Climate Change [34] | 40,000,000 | 25,728,395 | 39,567,031 | 8,029,516 | 5 | 7,507,050 |
| Trump Tweet Ids [35] | 40,202,199 | 16,690,791 | 40,202,199 | 9,408,459 | 5 | 15,233 |
| Health Care [36] | 254,971,894 | 79,348,847 | 132,234,469 | 22,762,224 | 30 | 58,003 |
| 2018 Congregational Election [37] | 60,689,821 | 33,257,138 | 60,689,821 | 9,792,467 | 7 | 30,601 |
| News Outlets [38] | 110,656,738 | 103,811,445 | 106,453,919 | 91,026,264 | 13 | 645,041 |
| Women's March [39] | 14,478,518 | 7,061,577 | 14,478,518 | 1,286,113 | 2 | 1,568 |
| US Govt Ids [40] | 9,673,959 | 9,085,817 | 9,599,813 | 6,933,491 | 1 | 111,892 |
| End of Term [41] | 5,655,632 | 5,288,040 | 5,649,948 | 4,116,967 | 1 | 50,713 |
| Nipsey Tweets [42] | 11,642,103 | 6,944,028 | 11,631,700 | 1,307,212 | 1 | 181 |
| Winter Olympics [43] | 13,816,206 | 8,336,254 | 13,763,243 | 1,530,613 | 2 | 638 |
| Dallas Shooting [44] | 7,146,993 | 3,683,170 | 7,146,993 | 1,224,715 | 1 | 2,249 |
| Charlottesville [45] | 3,015,437 | 1,517,338 | 3,011,996 | 327,856 | 0 | 161 |
| Twitter-Events-2012-2016 [46] | 147,055,035 | 80,675,871 | 147,055,035 | 35,454,578 | 17 | 856,543 |
| 115th U.S. Congress Tweet Ids [47] | 2,041,399 | 1,919,544 | 2,041,399 | 1,528,001 | 0 | 22,024 |
| Immigration Exec Order [48] | 16,875,766 | 7,108,723 | 16,809,653 | 2,088,736 | 2 | 2,690 |
| Irish news English tweets [49] | 198,725,860 | 100,359,505 | 198,725,860 | 45,924,135 | 23 | 371,543 |
| Black Lives Matter [50] | 17,292,130 | 6,460,739 | 17,231,264 | 2,527,358 | 2 | 3,272 |
| 2020 Presidential Election [51] | 802,029,566 | 366,187,559 | 702,684,385 | 143,239,345 | 93 | 459,062 |
| Tweets to Donald Trump [52] | 583,890,932 | 227,909,402 | 362,464,578 | 175,277,501 | 68 | 613,307 |
| Total | 2,905,714,184 | 1,357,409,820 | 2,410,671,168 | 690,920,817 | 336 | 11,718,796 |

## 3 Silver Standard Dataset Curation

To curate the silver standard dataset, we combined data from from Twitter Stream and Publicly available dataset. From Table 3 we can observe that there are natural disaster tweets in datasets which are not directly relevant to Natural disasters application. The primary reason to use publicly available datasets is to demonstrate the presence of significant signals in other datasets which can be utilized for this application. While we collected data from several datasets and the Twitter stream, there are several datasets which overlap with each other. For example, hurricane Dorian tweets were released by two different outlets. However, there are several similar tweets since most studies that collect tweets use the same endpoint. Combining our data filtered from the Twitter stream and hydrated datasets, we obtained 11,718,796 tweets. However, to avoid bias, we removed duplicate tweets from the combined data and retained **7,148,739** tweets. Additionally, there were some terms in our list which can be attributed to other events and not just natural disasters. For example, "death toll" is a term which is often associated with epidemics and mass shootings. So in order to further filter the tweets, we only retained tweets if they contain the term - "hurricane", "floods", "earthquake" and "quake". This additional filtering is to denoise the dataset. All the tweets are preprocessed by removing whitespaces, emojis, emoticons, urls and lower-caing the text. The silver standard dataset contains **846,927** tweets. These tweets contain tweets from three different types of natural disasters, i.e hurricanes, earthquakes, and floods. Listed below are a few samples of the preprocessed tweets obtained through the heuristic. The tweet text has been paraphrased since tweet texts cannot be published. To summarize, we created a heuristic by generating bigrams from existing natural disasters datasets and identified the relevancy of the heuristic. To denoise the silver standard dataset, which increases the quality of the dataset, we added an additional check to retain tweets which contain tweets natural disaster tweets. Our heuristic of 155 terms could filter 846,927 natural disaster tweets which from is termed as silver standard dataset. The heuristic does not contain any of the labels from the gold standard dataset and we did not use any annotated dataset to create the heuristic.

1. "flood waters as deep as four feet close roads in many southern wisconsin counties"
2. "number of terengganu flood victims swells to 2,000"
3. "taiwan earthquake: buildings tilt on sides after at least four killed and scores missing amid rescue operation"
4. "death toll rises further, hundreds left homeless as hurricane irma devastates the caribbean"

## 4 Methods

To train the machine learning models, we used the silver standard dataset tweets as positive tweets. To balance the data, we added an equal number of non natural disaster tweets. Since the language on Twitter changes every year and to avoid language bias, we obtained an equal number of non natural disaster tweets from the publicly available datasets and Twitter Stream. A non natural disaster tweet is a tweet which does not match with any of our terms in the heuristic. We emphasize that we have not manually annotated any tweets, or select from any manually annotated tweets from others, in the silver standard dataset. We used a publicly available, manually annotated dataset to test our methods.

Table 4: Machine learning models description

| Model Name | Model base / embeddings | Model Description |
|---|---|---|
| BERTweet (BT) | bertweet-base | 12-layer, 768-hidden, 12-heads, 135M parameters |
| CNN | N/A | Adam Optimizer, Relu Activation Glove Embedding model [60] |
| Disaster-Bert-Tweet (DBT) | disaster-tweet-bert | 12-layer,768-hidden,12-heads |
| Naive Bayes (NB) | Multinomial Naive Bayes | default hyperparameters |
| Random Forest (RF) | N/A | criterion set to entropy and max_features set to auto |

Table 5: Mean F-measure of machine learning models; k represents thousand and M represents million

| #Size | BT | CNN | DBT | NB | RF |
|---|---|---|---|---|---|
| 100k | 0.6394 | 0.7862 | 0.58 | **0.9396** | 0.8679 |
| 200k | 0.5851 | 0.7663 | 0.5772 | **0.9392** | 0.8612 |
| 300k | 0.6772 | 0.7569 | 0.5605 | **0.9404** | 0.8556 |
| 500k | 0.5871 | 0.7242 | 0.5691 | **0.9415** | 0.852 |
| 1M | 0.6771 | 0.6948 | 0.6063 | **0.9439** | 0.8459 |

**Experimental Setup** To test the weak supervision approach, we experimented with several class balanced training sizes. We started from 100,000 samples and incrementally increased the training sample size to 1,000,000. For each training size, we selected the samples utilizing random sampling with a seed such that the results are reproducible. We experimented with 5 different training sizes (100,000, 200,000, 300,000, 500,000, 1,000,000), 5 machine learning models and 10 seeds for each training size. A total of 250 experiments were executed in this work. For each training size, we split the data into 80% (training dataset) and 20% (validation dataset). In a 1,000,000 training size, 500,000 samples are positive tweets (label 1) from silver standard dataset and 500,000 samples are negative samples (label 0).

**Machine learning models** We experimented with 2 classical models and 3 deep learning models. We utilized Scikit-Learn's [54] implementation of Naive Bayes (NB) and Random Forest (RF) classifiers. Scikit-learn's TF-IDF vectorizer was used to convert raw tweet text to TF-IDF features and return the document-term matrix which is sent to the model. On the deep learning models front, we utilized CNN model, and 2 transformer models i.e BertTweet (BT) [55] which is a pre-trained language model for English Tweets and Disaster-Tweet-Bert (DT) [56] models trained on disaster tweets. We used keras implementation [57] for the CNN models and Simple Transformers python library [58] for BertTweet and Diaster-Tweet-Bert which seamlessly utilizes the Hugging Face pre-trained models [59]. The following table 4 represents the model description for the machine learning models.

**Results** In order to evaluate the results, we used the following metrics: Precision (P), Recall (R), F-Measure (F) and Accuracy (A). Since we used 10 different seeds for each training size, we computed the average of all the experiments in each training size and presented the consolidated results. Table 5 presents the mean of F-Measure for all the models across all training sizes and Figure 2 depicts the progression of F-Measure. Each training size in Table 5 contains class balanced training samples i.e in 100k size there are 50,000 natural disaster tweets and 50,000 non natural disaster tweets.

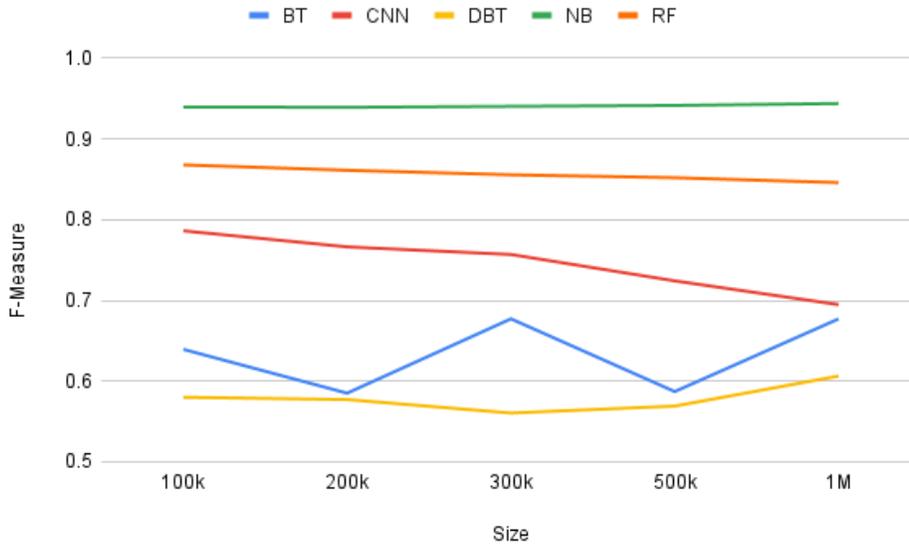

Figure 2: Mean of F-measure for machine learning learning models ; k represents thousand and M represents million

From table 4 we can observe that the models trained using the silver standard dataset can efficiently identify the gold standard dataset. We utilized existing models to demonstrate that this approach can be extended both classical and deep learning models. Since we did not optimize the hyperparameters, the results might not be state of the art. Yet, these results are equally competent when compared to the models trained on smaller gold standard datasets. While we built a heuristic to identify tweets from three different disasters, we utilized a binary classification instead of a multi-class classification to demonstrate the methodology. Multi-class classification is out of scope of this work. The primary reason for demonstrating a binary classification is to demonstrate that the silver standard dataset which contains tweets from several natural disasters is able to identify the natural disaster tweets from the gold standard. We only have different sub groups to demonstrate the different types of natural disasters.

BERTweet model was trained on 850 Million English tweets and Disasterbert-tweet was trained on disaster tweets which include not only natural disasters but also disasters like road accidents. These add significant noise to the model. In weak supervision, as the noise increases, the performance of the model tends to decrease. Further, when using silver standard data, it is important to experiment with multiple models to identify which model works best for an application. Further, We would like to emphasize that this is the first application to utilize weak supervision for natural disasters research.

## 5 Dataset Deliverables

To encourage reproducible research, we release our silver standard dataset and code. Due to Twitter's terms of service, tweet text cannot be shared. Therefore, tweet ids are publicly made available using Zenodo. The whole methodology can be reproduced using the deliverables.To reproduce this research and use the data, researchers must obtain a Twitter developer account to hydrate the dataset. The released dataset adheres with FAIR principles [61] in the following ways: The dataset is Findable as it can be accessed with a persistent DOI (Digital Object Identifier) in Zenodo. The dataset is Accessible through the DOI. The dataset contains only tweet identifiers as tweet text cannot be shared as per Twitter's terms of Service. However, tweets might be deleted either by Twitter or the user. In such cases, we can share the data on request while adhering to the Twitter data sharing policy. The tweet identifier can be hydrated to a tweet json object using tools like Social Media Toolkit [23] or Twarc [62]. The hydrated tweets are json objects which are derived from JavaScript object notation syntax. JSON is a universally accepted format thus supporting Interoperability. This dataset is released with Creative Commons Attribution 4.0 International for Reusability. The dataset is available at https://doi.org/10.5281/zenodo.6628961

**Social Impact** We believe that there are no negative social impacts of this work. However, since the silver standard data is not verified by a human, there might be "noise" in the data which might not be a complete fit to address intricate problems in the domain. Researchers can definitely alter the heuristic to their problem statement and refine the silver standard data.

**Limitations** There are few limitations to the dataset. We include the code to reproduce the gold standard dataset however, we do not include the tweet ids of the gold standard dataset since we did not curate the set. Researchers might need to reach out to the authors of the gold standard dataset if they remove their publicly available dataset.

To reproduce the method to obtain the dataset, researchers must hydrate publicly available dataset and also obtain tweets from the Twitter stream which might not be possible because it is difficult to retain tweets from an end point after each day. In the publicly available datasets, tweets might be deleted by the user/ Twitter. Additionally a lot of computational power and storage is required to hydrate, store and process the tweets which is difficult to do on a work station with small capacity. We encourage researchers to utilize the silver standard dataset and heuristic for natural disasters research

## 6 Future Work

The proposition of this work is to utilize social media data and apply a weak supervision approach to obtain training data and characterize three different types of natural disasters. There are several directions in which this work can be extended. Firstly, the experiments in this paper demonstrate promising results when tested on a gold standard. However, an empirical experimentation and evaluation would further strength the validity of the dataset. Secondly, we would we would explore the possibilities of using weak supervision on class imbalanced data as well as multi-class data instead of a binary classification. Additionally, we would like to experiment to see how well the models generalize for other natural disasters like Typhoons and Tsunamis. In future, we would like to train a BERT model on the silver standard dataset and release it for the scientific community.

## 7 Conclusion

In this work, we curated a silver standard dataset using weak supervision to characterize three different types of natural disasters. We believe that we cannot train models with a very limited amount of manually annotated tweets, but we can use the theory of noisy labeling to create more robust models with silver standard datasets. We collected over 7 billion tweets and filtered 846,927 tweets using a heuristic. We trained several machine learning models to experiment with the weak supervision approach and our results validate our dataset. The silver standard dataset can certainly be utilized for natural disaster research as a training set for supervised models. Further, this approach can definitely be reused and extended to several other domains by changing the heuristic and the filtering mechanisms.